# Visual Attention Is Beyond One Single Saliency Map

Jian Li
NUDT
Changsha P.R.Chian
lijian.nudt@gmail.com

## Abstract

*Of later years, numerous bottom-up attention models have been proposed on different assumptions. However, the produced saliency maps may be different from each other even from the same input image. We also observe that human fixation map varies across time greatly. When people freely view an image, they tend to allocate attention at salient regions of large scale at first, and then search more and more detailed regions. In this paper, we argue that, for one input image visual attention cannot be described by only one single saliency map, and this mechanism should be modeled as a dynamic process. Under the frequency domain paradigm, we proposed a global inhibition model to mimic this process by suppressing the non-saliency in the input image; we also show that the dynamic process is influenced by one parameter in the frequency domain. Experiments illustrate that the proposed model is capable of predicting human dynamic fixation distribution.*

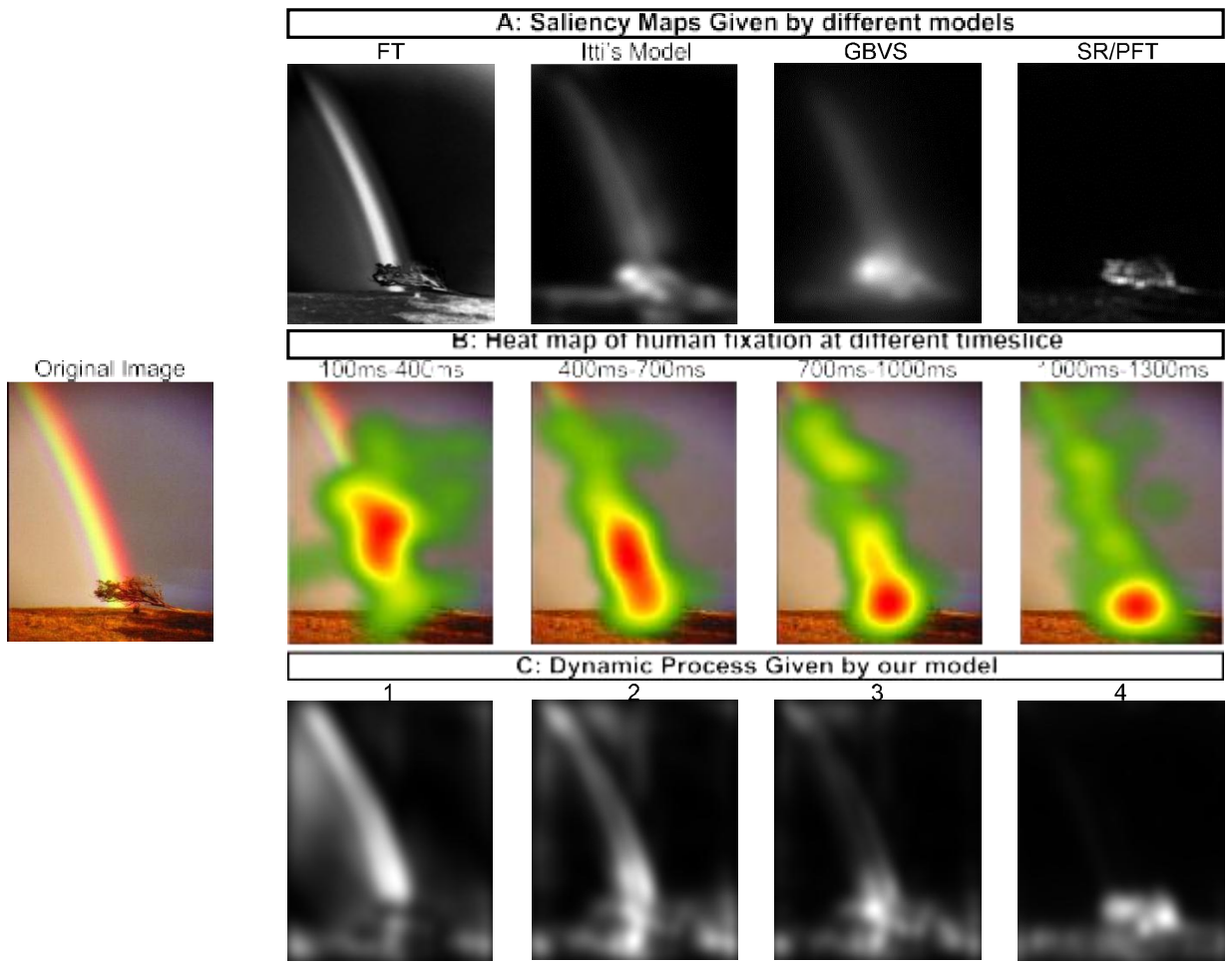

Figure 1. Visual selective attention is a dynamic process. **A:** saliency maps produced by frequency-tuned (FT) model[1], Itti's Model[2], GBVS model[3]and SR[4]/PFT[5, 6] model; **B:** the fixation maps are diverse at different time-slice; **C:** the proposed model produces a series of saliency maps by adjusting a parameter in the frequency domain.

## 1. Introduction

Visual saliency has received extensive attention in both psychology [7, 8, 9] and computer vision [2, 10, 11, 12,13, 14, 3] domain, and the goal of it is to reveal the mechanisms of visual attention and fixation behavior of primate visual systems. Objects in scenes viewed by human visual system are considered to compete with each other to distribute our attention to a subset selectively [15]. By suppressing each other, objects will influence how they are viewed in the visual field, consequently many of them are inhibited, while those are not will predominate in the visual cortex to cause a focus of attention [9]. Two different processes influence visual attention: one is top-down, which depends on the task at hand; the other is bottom-up, which is driven by the input image [16]. In this paper top-down is not considered.

There are many computational models in literature, and several of them utilize the local information. Itti and Koch's saliency model[2]is the exemplar for saliency detection and is consistently used for comparison in literatures. Gao *et al.* [17] proposed a bottom-up saliency model by using Kullback-CLeibler (KL) divergence to measure the feature difference between a location and its surrounding area. Recently, several new models compute saliency by using global information. In [1], the input color image is represented in the Lab space (an opponent color space), and then the saliency value of each location is defined as the difference between the Lab pixel value and the mean Lab value. Harel *et al.* [3] proposed a graph-based solution which uses local computation to obtain a saliency map, which is everywhere dependent on global information. Hou and Zhang [4] proposed a Fourier Transform-based saliency model, called Spectrum Residual (SR). Successively, the Phase spectrum of Fourier Transform (PFT) was presented, which achieved nearly the same performance of SR [6]. In recent years, with the development of the eye tracking technology, which can record the locations of human attention fixations, many researchers pay their attention to the investigation and the simulation of gaze path., e.g. using the "winner-take-all" and "inhibition of return" strategies[2].

In those models, only one single saliency map is produced for an image on different assumptions. Such a saliency map is considered as a probability map, and the saliency value at each location indicates the chances of how likely people paying attention there. Surprisingly, the saliency maps produced by those models are always different from each other greatly even for the same input image, as shown in row A of Fig.1. But which one is correct? We consider visual attention is a dynamic process, and the models mentioned above are just simulating different stages of the whole process. We think all of them are reasonable saliency results.

Attention selection facilities primates to perceive targets in the environment. In visual attention at least two properties of targets should be counted: one is their **location** and the other is the **size**. In the past, researchers only paid attention to the "location", while the "size" of the salient regions was not considered enough.

We argue that, the property–"the size of the target" also plays an important role in human selective attention, which is to say, what people can perceive at one moment is both driven by the information of the input image and the focus length of his/her eyes at that moment. When viewing a picture or scene, people usually need to adjust their *focus length* to search targets of different sizes. In free viewing, people tend to allocate their attention at those broad salient regions (e.g. near objects, which are of larger *saliency scale* ) at first, then they will transfer attention to smaller salient regions (e.g. remote objects or other local informative locations, which are of smaller *saliency scale*), as shown in row B of Fig.1. What people see at different moments vary greatly, and these perceived targets/regions may overlap with each other. Hence, it's very difficult to describe where people pay attention by using only one single saliency map. Employing the frequency domain paradigm [4], we proposed a global inhibition model to mimic this process by suppressing the **non-saliency** of the input image; we also show that such a dynamic process is influenced by one parameter in the frequency domain. Experiments illustrate that the proposed model is capable of predicting the human dynamic fixation distribution.

The paper is organized as follows: section 2 discusses the dynamic feature of human fixation distribution; section 3 investigates the modeling of the dynamic attention selection process with the proposed model; section 4 discusses experimental results; concluding remarks and possible extensions of this work are shown in section 5.

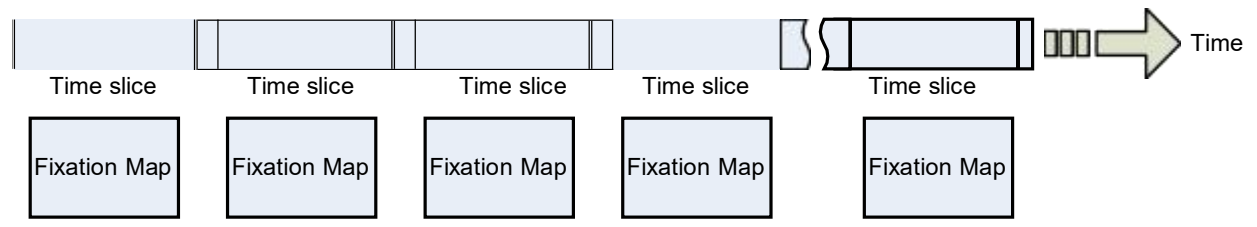

Figure 2. The fixation sequence comprises several fixation maps.

## 2. The dynamic feature of human attention

In our experiment, we use an eye tracker to collect human fixation data. Each stimuli (an image) is presented on screen for a certain period, during which the locations of participants' fixations are recorded (at 60Hz)[1]. Instead of creating only one fixation map with altogether the fixation data on one stimulus, we divide the period into several *time slices* and create a fixation map for each corresponding time slice with the data of intervals. Thus we have fixation map sequences for each image, as shown in Fig. 2.

From the fixation sequences of the stimulus, we find some interesting features of human attention.

**1) Attend to larger salient regions at first**.

Fig.3 presents two stimulus in vertical orders. In each stimulus, there are two disks of the same color, but of different sizes. In the upper stimuli, the larger disk is placed in the right side. In the lower one, the larger disk is put in the left side. Two mirrored images are presented to eliminate the potential reading preference (e.g. from left to right, or from right to left). From both of the two stimuli, we find that people tend to allocate attention at the larger disk rather than the smaller disk at first sight. For the upper stimuli, during 300ms-800ms, more than 75% fixations are allocated on the larger disk. In the next 1000 ms, the attention is transferred from the larger disk to the smaller one. This occurs for the second stimuli, too.

This phenomenon also occurs for natural images. From the obtained fixation data, we observe that people tend to perceive targets of larger size at first when an image is presented, then to smaller targets or regions with more details. As shown in row B Fig.1, most of participants pay attention to the rainbow[2] at first 400ms[3], and then attention will be transferred to the smaller target (the tree). This is because people like to search targets generally from the scene (with small eye focus length), and then search smaller targets or one part of large target (corresponding to large eye focus length). The explanation for this could be: 1) During the attention process, the focus length of our eye is adjusting from short to long continuously; 2) such a coarse-to-fine process

[1]for more information about the setup of this experiment, please refer to: http://www.cim.mcgill.ca/ lijian

[2]When people view the rainbow (which takes a large region in the image), the focus length of their eyes are very small, and they needn't move their gaze (fixation location) to perceive the whole of the rainbow.

[3]We find that, the fixation locations of the first 100ms of the presented image are always affected by the previous image, so we don't count this first 100ms.

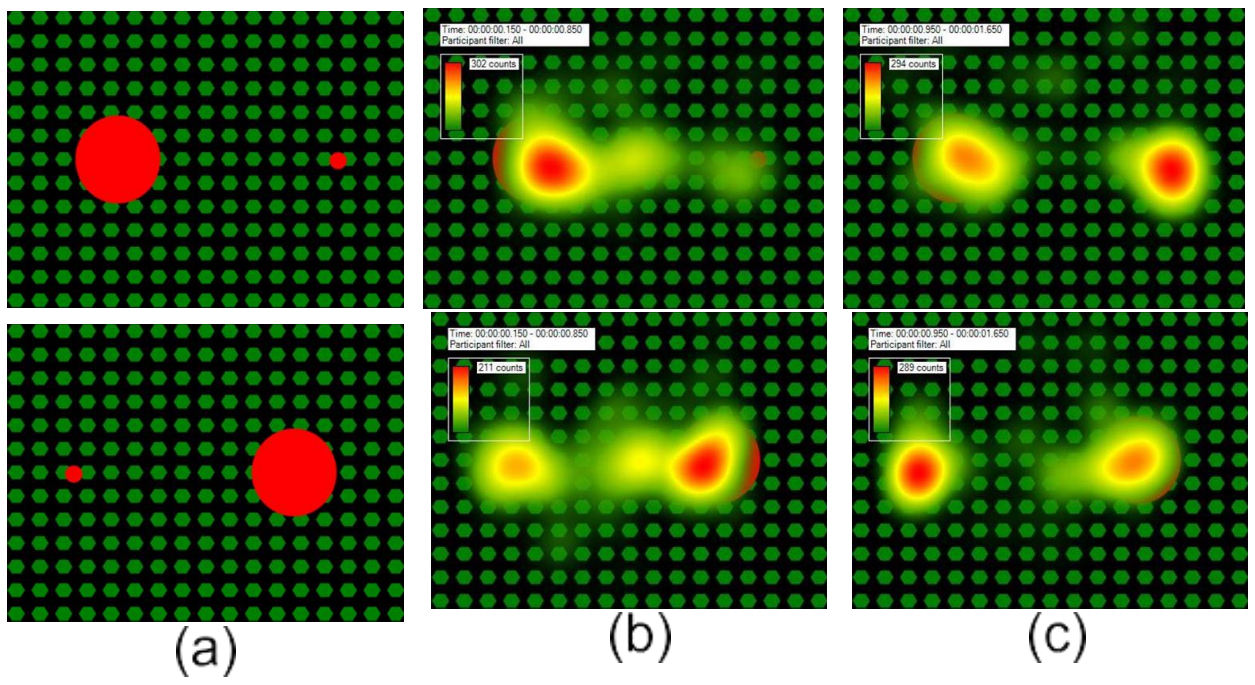

Figure 3. Human tend to allocate attention to larger salient regions at first. Column (a) shows two presented stimulus (the bottom one is the mirror of the upper one); column (b) shows the corresponding human fixation maps recorded at earlier period; while column (c) shows the human fixation maps at a later period.

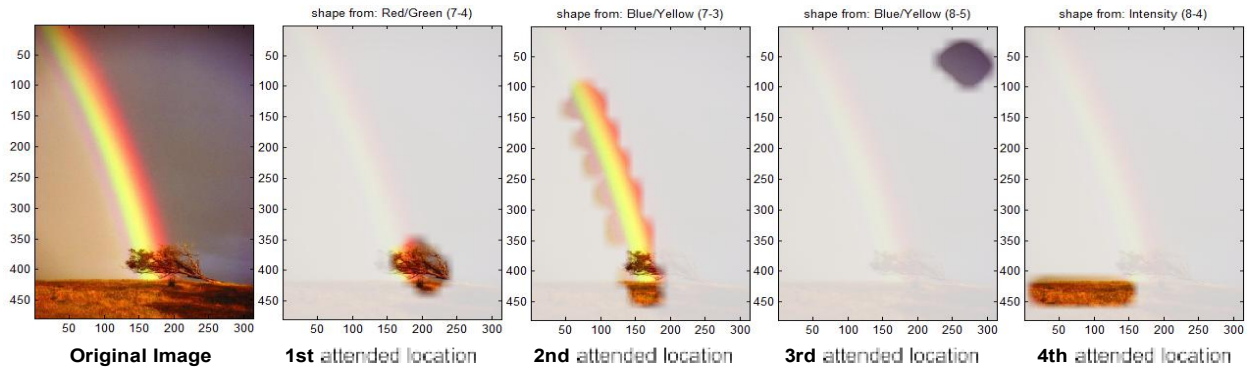

Figure 4. The *inhibition of return* strategy cannot always correctly predict human searching path.

helps people to seize the gist of the scene/image rapidly.

**2) The inhibition of return strategy cannot always predict human searching path correctly.**

Based on the discussions above, we consider human attention behavior is a dynamic process. However, do we really need to use a series of saliency maps to represent this dynamic process? Could we simulate this dynamic process by one single saliency map combining with the "inhibition of return" strategy[2, 6] ? The question is the inhibition of return strategy cannot predict human searching path correctly every time, as shown in Fig.4. As we know, in the inhibition of return strategy, the gaze path is determined by the local maximum of the saliency map, but the dynamic feature mentioned above is overlooked., According to the inhibition of return, the first attended location is the tree, not the rainbow, as shown in Fig.4. This is because the local maximum of the "tree" is larger than that of the rainbow[4].

**The multi-scale strategy in spatial domain cannot correctly predict the dynamic attention**

Someone may argue that the dynamic process can be simulated by any model by changing the scales of input image. In[4], it is argued that detecting salient regions of different sizes can be realized by changing the scales of the input image (it equals the different layers in the pyramid). However, in this paper, we will show that such a dynamic process cannot be simulated this way. In the experiment, we set up an image pyramid for a natural image, and created several saliency map sequences by some existing models, but the result shows the multi-scale strategy in spatial domain cannot be employed to simulate the dynamic process of human attention, as shown in Fig.5;

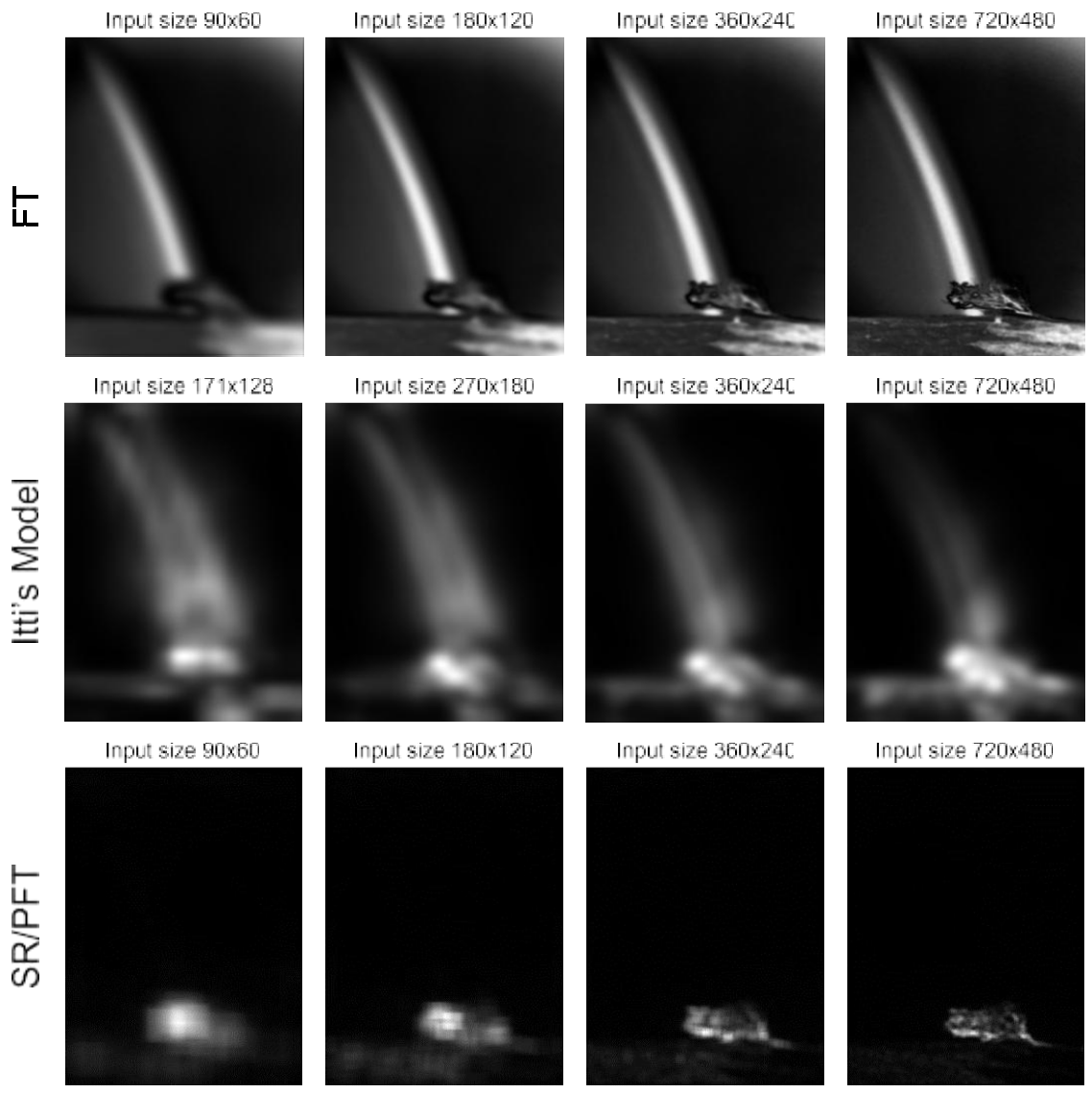

Figure 5. The *multi-scale* strategy in spatial domain cannot be employed to simulate the dynamic process of human attention.

In next section, we will discuss our proposed model. We will illustrate the dynamic process of human attention by using a global inhibition mechanism which is based on the scale space analysis in the frequency domain.

## 3. Selective attention by suppressing mutual inhibition parts

Many attention models were proposed, which invariably then require the detection of salient *regions*. These regions are described as *distinctive* or *irregular* patterns, which possess a distinct feature distribution when compared with the rest of the image. In this paper, instead of searching for these irregular patterns, we model these so-called common patterns that mutually inhabit each other, and do not attract much attention by our visual system. We refer to these patterns as being *nonsalient*.

### 3.1. Suppressing repeating patterns for saliency pop out

In the proposed model, we assume that a natural image consists of several salient and many so-called regular re-gions. All of these entities (whether distinct or not) may be considered as visual stimuli that compete for attention in the visual cortex. In this regard, it has been shown that nearby neurons constituting receptive fields in the visual cortex mutually inhibit each other and interact competitively [9]. As shown in Fig. 6, if we divide the image into many patches (at a particular scale), we find that, some are distinctive, while others are quite similar to each other. Fig. 6(a) shows the collection of patches from the natural image left. We observe that several patterns appear many times (e.g., blue sky and grassy patches). We refer to these regular patches as *repeated patterns*, which correspond to *non-saliency*.

[4] The dynamic feature of human attention is quite complex. Besides the coarse-to-fine strategy, it can also be affected by the center-bias effect, top-down influence, etc. Here we only investigate the influence of coarse-to-fine process.

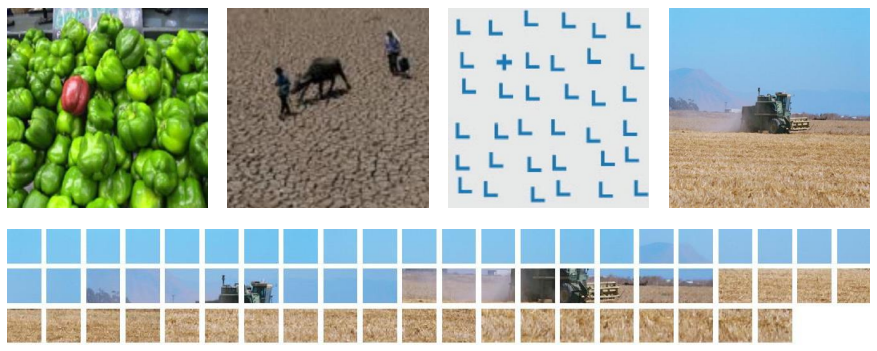

Figure 6. Regular (here also called common or repeated) and anomalous patterns. Top, four natural images; bottom: Collection of fragments from the last image.

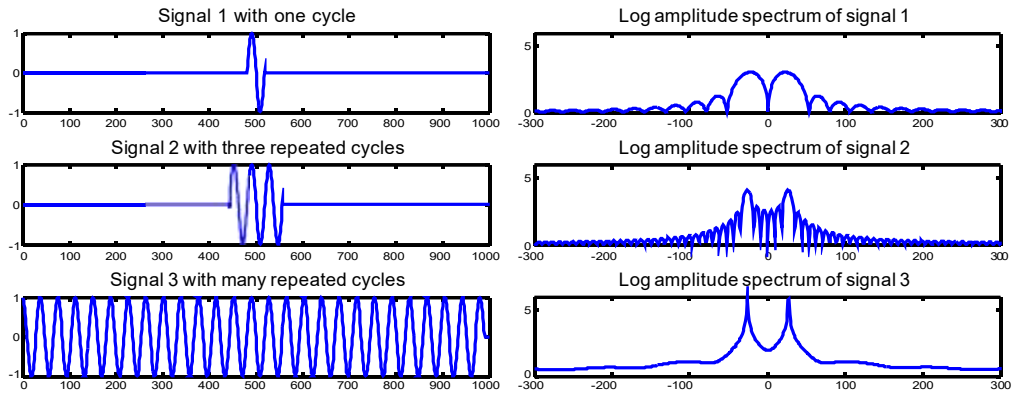

Figure 7. Repeated patterns lead to sharp spikes. Left: Signals with different number of repeated cycles; right: Corresponding amplitude spectrum in frequency domain.

Clearly, the primate visual system is more sensitive to distinctive rather than repeated patterns. Furthermore, the latter are very diverse. For example, consider the top row of Fig.6. These exhibit several different examples of repeated patterns at different scales (including at the scale of a single pixel for the uniform areas): grassy/sky patches (image 4), similar objects (image 1), road patches of the same color and texture (image 2), the 'L's (image 3), and so on. In this paper, we model these repeated patterns and then suppress them, thereby producing the pop-out of the salient regions.

## 3.2. Spikes in the amplitude spectrum correspond to repeated patterns

It has been argued [4] that the so-called spectrum residual corresponds to saliency, while contradictorily in [5], the amplitude information was totally abandoned. However, in this paper, we will illustrate that the amplitude spectrum also contains important information corresponding to saliency and non-saliency. To be more precise, the spikes in the amplitude spectrum turn out to correspond to repeated patterns, which should be suppressed for saliency detection.

For convenience, we take a 1-D periodic signal $f(t)$ as an example. Suppose that it can be represented by $f(t) = \sum_{n=-\infty}^{\infty} F(n) e^{jn\omega_1 t}$, where $F_n = \frac{1}{T}\int_{-T/2}^{T/2} f(t) e^{-jn\omega_1 t} dt$. Then the Fourier transform is given by:

$$F(\omega) = 2\pi \sum_{n=-\infty}^{\infty} F(n)\delta(\omega - n\omega_1). \tag{1}$$

From (1), we can conclude that the spectrum of a periodic signal (repeated cycles) is a set of impulse functions (spikes). We note that this is based on the assumption that the periodic signal is infinite. Therefore, given a more realistic finite length periodic signal, the shape of the spectrum will obviously be different but not degraded greatly.

Fig. 7 provides an example of this. Fig. 7 (a) shows three signals with a different number of repeated patterns (cycles) while Fig.7 (b) shows the corresponding amplitude spectrum. We observe that the larger the number of repeated cycles, the sharper the spectrum. In order to quantitatively analyze this notion, we define the *sharpness* of an amplitude spectrum $X$. Suppose that we smooth the amplitude spectrum, containing several spikes, using a low-pass filter. Then we observe that the sharper the original spike, the more its peak height will be reduced. Therefore the *sharpness* of $X$ can be defined as follows: $P(X) = \| X - X * h \|_\infty$ where $h$ is a Gaussian kernel with fixed scale $\sigma$. As shown in row 3 of Fig. 7, repeated patterns produce a sharp spike in the amplitude spectrum. Besides a sinusoid, other repeated signals also have this characteristic.

Suppose there is one salient part that is embedded in a finite length periodic signal (see the original signals in Fig. 8). We will illustrate that this salient interval will not largely influence the spikes in the spectrum. That is to say, 1) The spikes will remain even though a salient part is embedded in the signal; 2) The embedded salient part will not lead to very sharp spikes in the amplitude spectrum. The signal to be analyzed is defined as follows:

$$f(t) = g(t) + g_0(t) + s(t), \tag{2}$$

where

$$g(t) = \begin{cases} p(t) & \text{if } t \in (0, L) \\ 0 & \text{otherwise} \end{cases}, \tag{3}$$

$g_0(t) = -p(t) \cdot W(t)$, $s(t) = p_s(t) \cdot W(t)$; $s(t)$ is the salient part of $f(t)$, which for convenience is also defined as a portion of yet another periodic function $p_s(t)$; $p(t)$ and $p_s(t)$ are periodic functions with frequencies $f$ and $f_s$ respectively; $W(t)$ is a rectangular window function that equals 1 inside the interval $(t_0, t_0 + \sigma)$ and 0 elsewhere; we also suppose that $(t_0, t_0 + \sigma) \in (0, L)$ and $\sigma \quad L$. Thus the Fourier Transform of $f(t)$ can be represented as follows:

$$F(f)(\omega) = \int_{-\infty}^{\infty} f(t)e^{-j\omega t}dt = \int_{0}^{L} g(t)e^{-j\omega t}dt + \int_{t_0}^{t_0+\sigma} g_\sigma(t)e^{-j\omega t}dt + \int_{t_0}^{t_0+\sigma} s(t)e^{-j\omega t}dt. \quad (4)$$

From (4), the spectrum of $f(t)$ consists of three terms. We assume that $\sigma \quad L$. This implies that the first term has very sharp spikes in the amplitude spectrum as it contains many repeated patterns, while this is not true of the second and third terms. Consider $g_\sigma(t)$ as an example. $g_\sigma(t)$ is the point-wise product of a periodic signal $-p(t)$ and a rectangular window function $W(t)$. According to the convolution theorem, $(Fg_\sigma)(\omega)$ equals the convolution of $(p)(-\sigma F)$ with $F(W)(\omega)$. Since $F(W)(\omega) = \frac{2\sin(\sigma/2)}{\omega} e^{j\omega(t_0+\sigma/2)}$ is a low-pass filter, the spikes in the amplitude spectrum of $-(Fp)(\omega)$ will be greatly suppressed. That is to say, there are no sharp spikes in the second term. This also occurs for the third term. As discussed above, the sharpness of $F(f)(\omega)$ is mainly determined by $g(t)$, while the latter two terms in (4) do not make much contribution to the spikes in the spectrum. In other words, since the first term corresponds to repeated patterns (non-salient) which lead to spikes, they can be suppressed by smoothing the spikes in the amplitude spectrum of $F(f)(\omega)$.

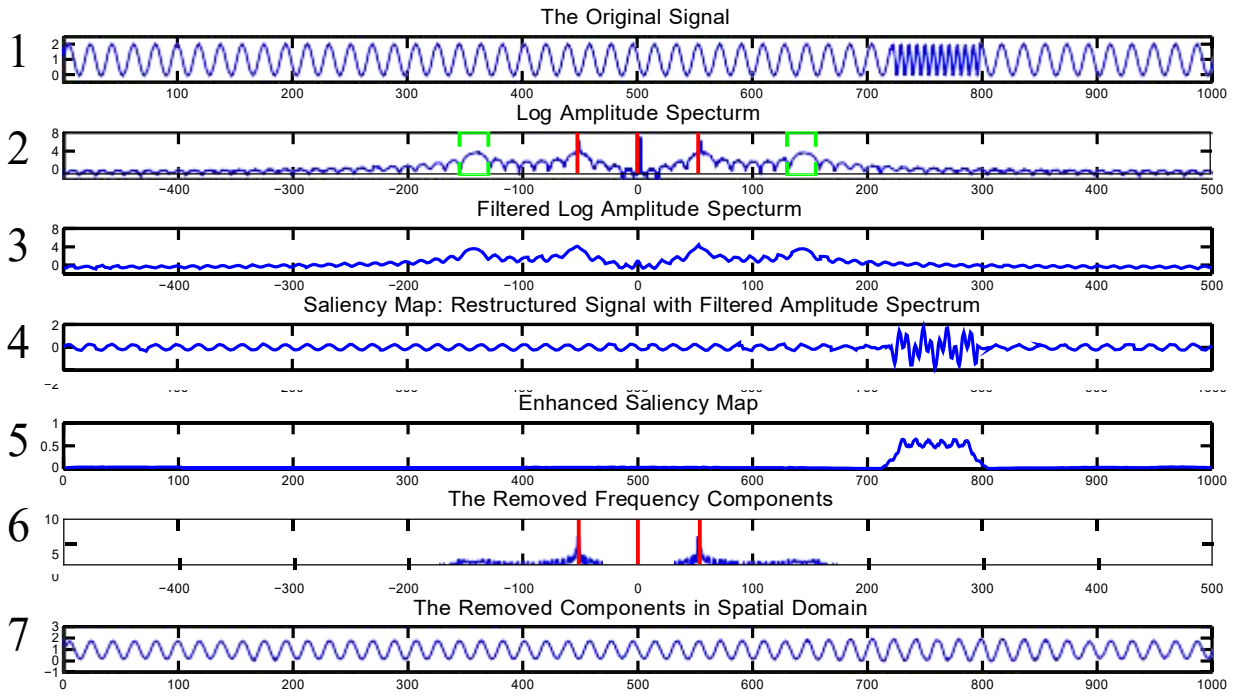

Figure 8. Suppression of repeated patterns by using spectrum filtering. Analyzing this process, it is clear that the larger the background, the sharper the spikes, leading to the suppression of the amplitude spectrum via filtering.

## 3.3. Suppressing Repeated Patterns Using Spectrum Filtering

A Gaussian kernel can be employed to suppress spikes in the amplitude spectrum as follows[5]:

$$A_S(u, v) = |F\{f(x, y)\}| > g, \quad (5)$$

where $g$ is a Gassian kernel with a scale $\sigma$ and $|F\{f\}|$ is the amplitude spectrum of a signal $f(x, y)$. The resulting smoothed amplitude spectrum $A_S(u, v)$ and the *original* phase spectrum are combined to produce the inverse Fourier Transform, which in turn, yields the saliency map:

$$S = F^{-1}\{A_S(u, v)e^{i \cdot P(u,v)}\}. \quad (6)$$

In order to improve the visual display of saliency, we define it hereafter as:

$$S = g > |F^{-1}\{A_S(u, v)e^{i \cdot P(u,v)}\}|^2. \quad (7)$$

Consider the very simple example shown in Fig. 8. The input signal (row 1) is periodic, but there is a short segment for which a different frequency signal is apparent. Note that the whole signal is superimposed on a constant value. The short segment is quite distinct from the background for human vision, so a saliency detector should be able to highlight it. Row 2 shows the amplitude spectrum: there are three very sharp spikes (labeled by solid boxes) which correspond to the constant at zero frequency plus two, which correspond to the periodic background. In addition, there are two rounded maxima (labeled by a dashed box) corresponding to the salient parts. The complete amplitude spectrum is then smoothed by a Gaussian kernel (row 3), and the signal is reconstructed in the spatial domain using the smoothed amplitude and original phase spectrum (row 4). It is clear that both the periodic background and the near zero-frequency components are largely suppressed while the salient segment is well preserved. Row 5 shows the (spatial domain) saliency map after enhancing the signal shown in row 4 using post-processing. We can further analyze this in the frequency domain, as shown in row 6, which illustrates the components actually removed by the previous operations. Here the eliminated frequency components are mainly the low frequencies near zero frequency, as well as the periodic background. Row 7 presents these removed components in the spatial domain (by measuring the difference between the original and reconstructed signals). Note that Row 6 indicates the frequency spectrum of the signal shown in row 7. Note that we perform the convolution (smoothing) discussed above in the frequency domain using only amplitude spectra and ignoring the phase[6]. This is very different from the process described in the *Convolution Theorem*[7].

[5]In the computer implementation of this, we found that suppressing spikes in the log amplitude spectrum rather than the amplitude spectrum yielded better results.

[6]This is done notwithstanding the fact that the Fourier Transform is actually always complex.

[7]We have argued above that convolution in the frequency domain of the amplitude spectrum with a low-pass filter is equivalent to an image saliency detector in the spatial domain. Ostensibly, this conclusion is similar to Convolution Theory. However, this is not true. As we know, there are two cases in Convolution Theory. Given an monochrome image $f(x, y)$ and a 2D Gaussian kernel $g(x, y)$, one case can be summarized as: $D = f \, y \, g \Leftrightarrow F\{D\} = F\{f\} \cdot F\{g\}$, which implies that

## 4. Modeling dynamic attention process using Spectrum Scale Space analysis

Repeated patterns can be suppressed by smoothing the amplitude spectrum at an appropriate scale. However, what does the scale $\sigma$ of the kernel in (5) indicate? In this section, we will show that such a parameter of that kernel controls the saliency scale, just like the adjustable focus length of our visual system. Hence the dynamic process of primate visual system can be simulated by performing scale space analysis in the frequency domain.

In this paper, we propose a Spectrum Scale-Space approach for handling amplitude spectra at different scales, yielding a one-parameter family of smoothed spectra which is parameterized by the scale of the Gaussian kernel. Given an amplitude spectrum, $A(u, v)$, of an image, the SSS is a family of derived signals $\Lambda(u, v; k)$ defined by the convolution of $A(u, v)$ with the series of Gaussian kernels:

$$g(u, v; k) = \frac{1}{\sqrt{2\pi}\, 2^{k-1} t_0} e^{-(u^2+v^2)/(2^{2k-1} t_0^2)}, \tag{8}$$

where $k$ is the scale parameter, $k = 1, \ldots, K$. $K$ is determined by the image size: $K = \lceil \log_2 \min\{X, Y\} \rceil + 1$, where $X$, $Y$ indicate the height and width of the image; $t_0 = 0.5$. Thus scale space is defined:

$$\Lambda(u, v; k) = (g(., .; k) * A)(u, v). \tag{9}$$

Fig.9 shows the workflow of the proposed model. The input image is firstly transformed into frequency domain by using Fourier Transform, then amplitude and phase spectra can be obtained; thus a spectrum scale space can be derived according to (9), see part C of Fig.9. From left to right, the scale of Gaussian kernel, which is used to blur the amplitude, is increasing. For each layer of the spectrum scale space, combining with the original phase spectrum, a saliency map can be produced by performing inverse Fourier Transform on the filtered amplitude spectrum. Thus, a sequence of saliency maps is produced. Part D of Fig.9 shows two examples of these saliency maps superimposed on one original image. This saliency map sequence mimics the dynamic process of primate attention system. Like what we sense in daily life, when a person appears before us, we pay attention to the whole body at first; then pay attention to his face, head and so on; and then pay attention to his eyes, ears, mouth or some even detailed parts. During this attention selection process, the focus length of our eyes is adjusting continuously. As what is shown in part D of Fig.9, we tend to firstly pay attention to the whole head and the hand regions of the man; a moment later, we tend to pay attention to the eye, ear and mouth of the man. Taking the hand region for an example, when we first glance at this region, we see the whole hand because it's quite salient compared with the background, however, we will not pay attention to more details at this moment; when we have watched this image for a longer period, our attention will be allocated to more detailed regions, like his raised thumb, which it's more distinctive compared with the other four fingers, hence attracting more attention. The proposed model can locate such a distinct region successfully.

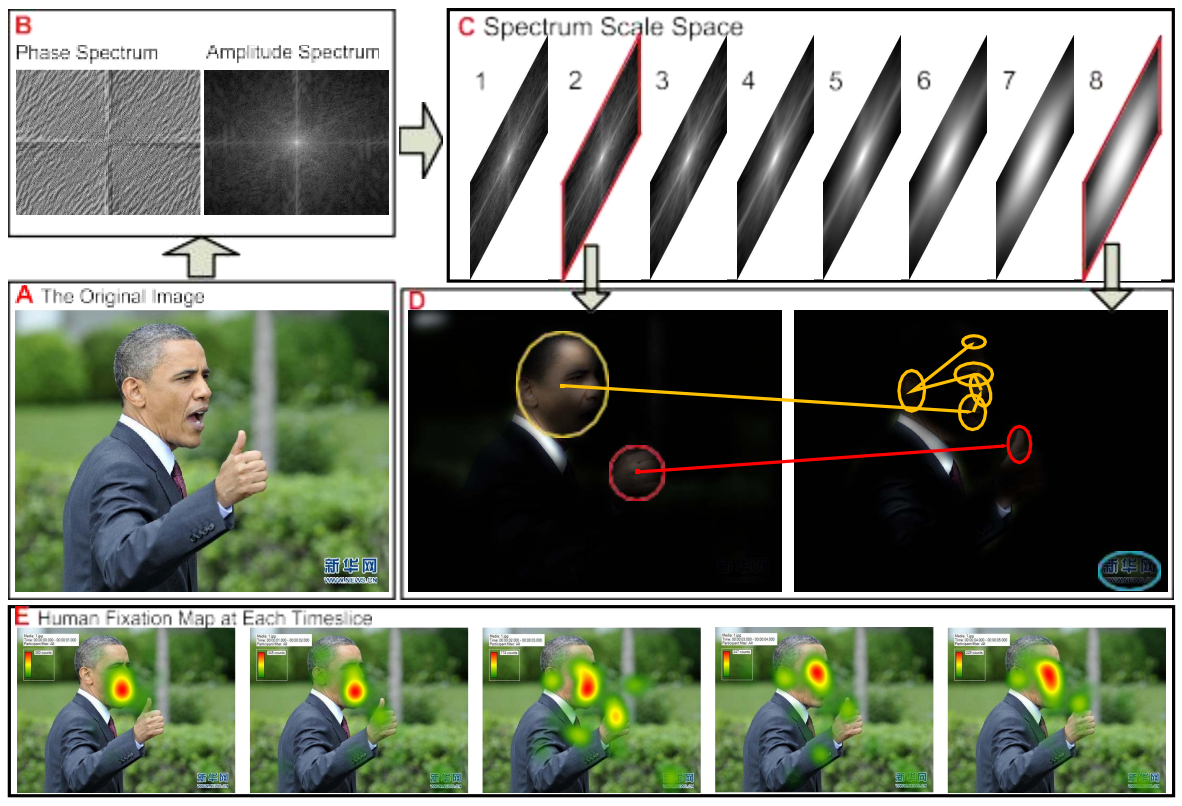

Figure 9. The work flow of the proposed model. Part A shows the original image; part B shows the amplitude and phase spectra; part C shows the **Spectrum Scale Space**, which is derived from the original amplitude spectrum convoluted by a series of Gaussian kernels; part D shows the two saliency maps (superimposed on the original image). These two saliency maps are obtained by performing inverse Fourier Transform on the corresponding filtered amplitude spectrum and the original phase spectrum. From these two saliency maps, we find that, the amplitude spectrum filtered by Gaussian kernel of smaller scale yields larger salient regions (larger saliency scale); while the spectrum filtered by Gaussian kernels of larger scale produces smaller salient regions (more detailed saliency).

Our conclusion is that, detecting salient regions of different sizes can be realized easily in frequency domain by suppressing *mutual inhibiting patterns* at different scales. This simulates the dynamic process of primate attention mechanism in computational way.

convolution in the spatial domain is equivalent to multiplication in frequency domain. The second case can be summarized as: $D = f \cdot g \Leftrightarrow F\{D\} = F\{f\} * F\{g\}$, This implies that convolution in the frequency domain is equivalent to multiplication in the spatial domain. It is worthwhile noting that both $F\{f\}$ and $F\{g\}$ are *complex matrices*. That is to say, it in general, convolution is performed between two *complex matrices*. Ostensibly, the proposed model is quite similar to case 2. Actually, they are totally different. The proposed model can be summarized as follows: $S(f) \Leftrightarrow |F\{S\}| = |F\{f\}| * |F\{g\}|$, where $S(f)$ represents the saliency in image $f(x, y)$ (Note that all saliency maps in this paper are actually given by $|S(f)|^2$, as done in [5, 4]). The right side of last equation implies convolving the amplitude spectrum $|F\{f\}|$ with a low pass Gaussian kernel (This is another form of (5). As we know, both g and $|F\{g\}|$ are low pass Gaussian kernels.) while both phase spectra remain unchanged. In summary, we can clearly observe the difference between the proposed model and the Convolution Theorem (case 2). The convolution in the proposed model is performed between two *real matrices*, while the convolution in case 2 of convolution theory is performed between two *complex matrices*.

## 5. Experiment

In this section, we will evaluate the proposed model both qualitatively and quantitatively. We apply a database to evaluate the proposed model.

In the first experiment, we choose three values $\sigma 1$, $\sigma 2$, $\sigma 3(\sigma 1 < \sigma 2 < \sigma 3)$ as the scales parameter of Gaussian kernels, and three saliency maps can be obtained for one image. Thus we can use the three saliency maps to simulate the dynamic attentional process. From discussion in section4, $\sigma 1$ corresponds to the larger scale saliency, and $\sigma 3$ corresponds to the smaller scale saliency. For each image, we choose three fixation maps, which correspond to three respective time slices $S_{t1}$, $S_{t2}$, $S_{t3}$, and $t1 < t2 < t3$. $ti$ is the midpoint of time slice $S_{ti}$. We also have compared the dynamic attention with the *inhibition of return*-based gaze path results[2]. Fig.10 shows the comparison results. Taking the left-bottom results as an example, in the original image, there are two salient objects, a near large yellow guidepost and a remote person riding on a bicycle. The second row of the results shows the saliency maps produced by our proposed model, in which the large guidepost pops out first; then the person emerge gradually; at last the person pops out completely, and the design in the guidepost is also attracting much attention at the same time. Our results are consistent with human fixation data (row 3). While, the "inhibition of return" detects the remote person at first, and then highlights the guidepost partially.

Similar to the qualitative experiments, we choose two values $\sigma 1$, $\sigma 2(\sigma 1 < \sigma 2)$ for the scales of the Gaussian kernel; for each value, the model produces one saliency map set, thus we have two saliency map sets, naming $\sigma 1$, $\sigma 2$ for short; for each stimuli, we choose two fixation maps corresponds to two different time slices $S_{t1}$, $S_{t2}$, and $t1 < t2$; thus we have two ground truth, naming $ttT1$, $ttT2$, respectively.

In the quantitative experiment, we perform cross-validation to evaluate the performance of the proposed model. According to the analysis in section4, if the saliency maps given by the proposed model is consistent with the fixation data, the $\sigma 1$ saliency map set are supposed to predict the $ttT1$ better than $\sigma 2$, while the $\sigma 2$ saliency map set should predict the $ttT2$ better than $\sigma 1$. Fig.11 shows the ROC curves for predicting the ground truth datasets.

## 6. Conclusions

In this paper, we consider attention selection as a dynamic process rather than that described by a single saliency map. The attention model proposed in this paper is based on two considerations. First, the mechanism of competing and mutual inhibition in primate visual system is taken into account. In this model, instead of modeling the salient regions using low level features, we model the non-saliency parts in the image, which will mutually inhibit each other. We argue that convolution of the amplitude spectrum in the frequency domain with a low-pass Gaussian kernel equals saliency detection in spatial domain. Second, we consider attention selection as a dynamic process, and such a process is embodied in the proposed model by performing scale space analysis in the frequency domain. We demonstrate experimentally that the proposed model is capable of predicting the human dynamic fixation distribution.

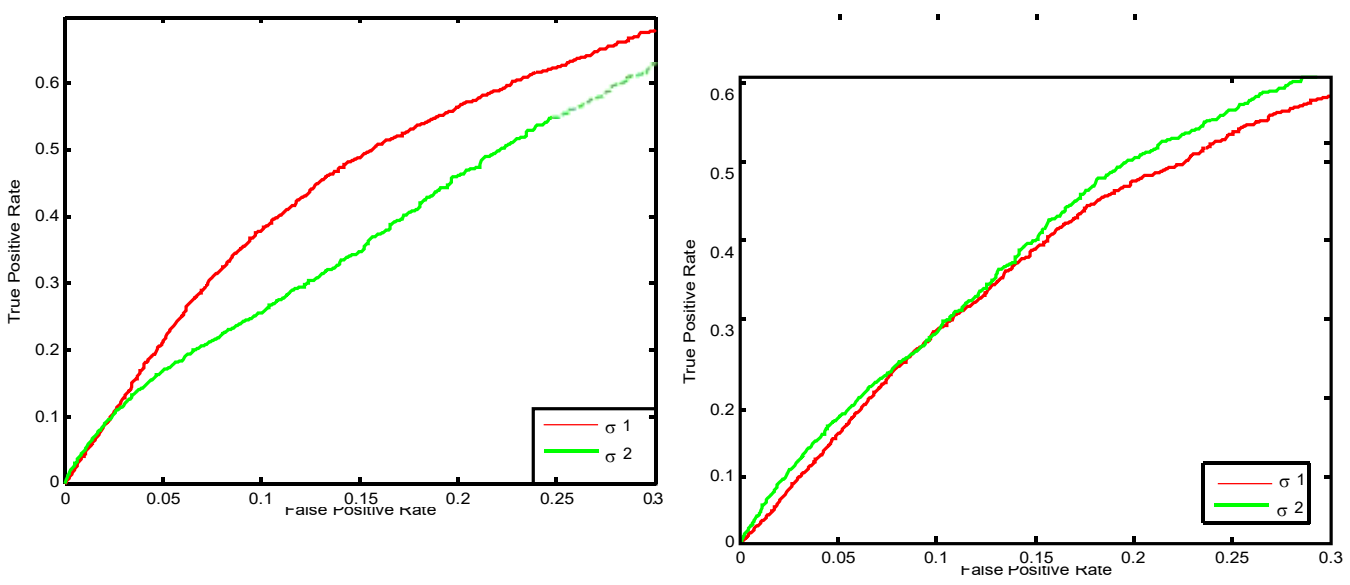

Figure 11. ROC curves of cross-validation. Left, $\sigma$ works well then $\sigma$ in predicting $ttT$ ; right, $\sigma$ works well then $\sigma$ in predicting $ttT$ ;

## 7. Acknowledgement

This work has been supported by National Natural Science Foundation of China (Grant No:91220301).

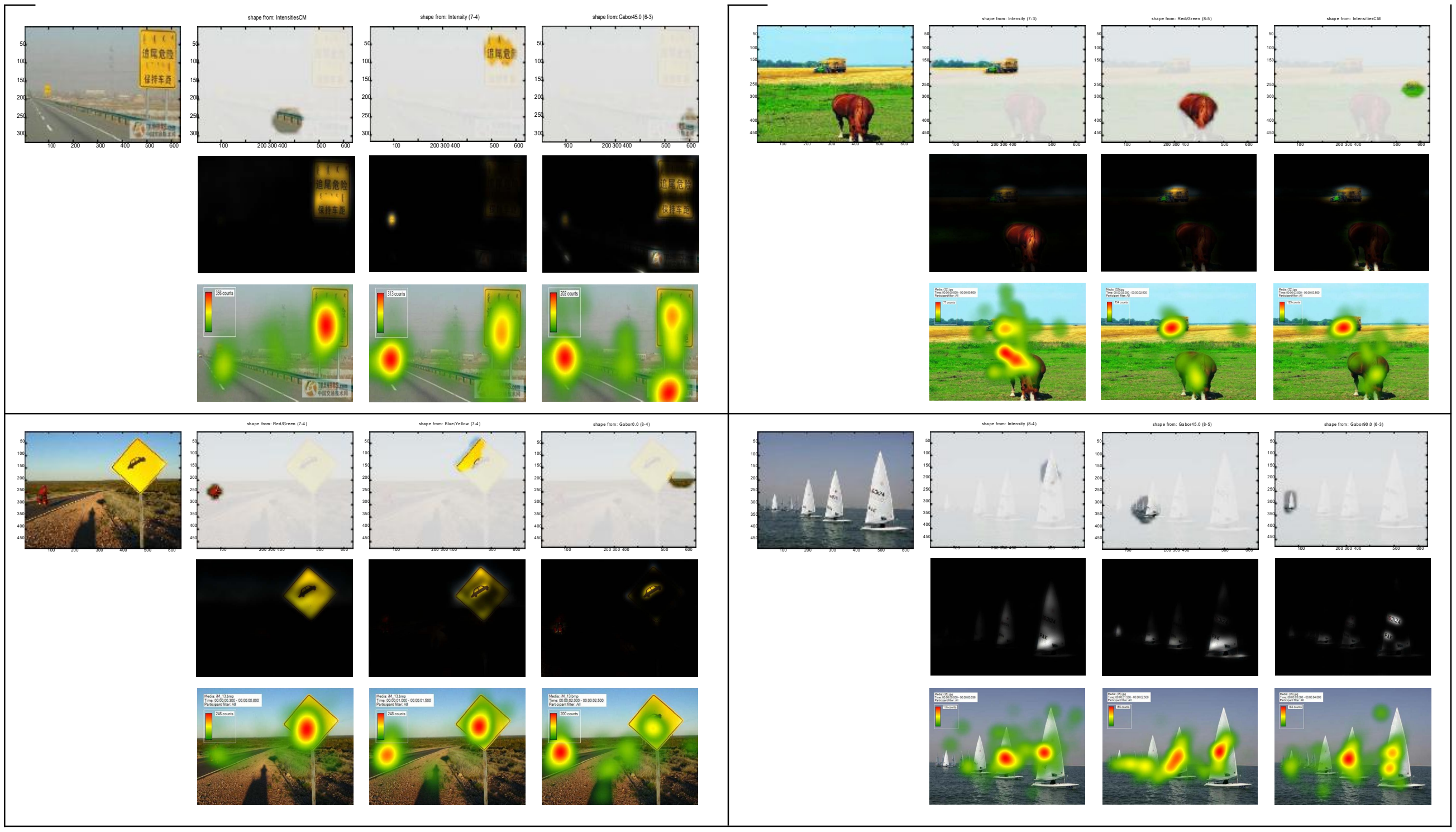

Figure 10. Experimental results on natural images. In each grid: the third row shows the fixation sequence recorded by the eye-tracker; the second row shows the saliency map sequence produced by the proposed model; the first row shows the attended regions in sequence[2].